\documentclass[10pt,letterpaper,twocolumn,english]{article}
\usepackage[T1]{fontenc}
\usepackage[latin9]{inputenc}
\usepackage{array}
\usepackage{amsmath}
\usepackage{amssymb}
\usepackage{graphicx}

\makeatletter

\DeclareRobustCommand{\greektext}{%
  \fontencoding{LGR}\selectfont\def\encodingdefault{LGR}}
\DeclareRobustCommand{\textgreek}[1]{\leavevmode{\greektext #1}}
\DeclareFontEncoding{LGR}{}{}
\DeclareTextSymbol{\~}{LGR}{126}
\newcommand{\lyxmathsym}[1]{\ifmmode\begingroup\def\b@ld{bold}
  \text{\ifx\math@version\b@ld\bfseries\fi#1}\endgroup\else#1\fi}

\let\SF@@footnote\footnote
\def\footnote{\ifx\protect\@typeset@protect
    \expandafter\SF@@footnote
  \else
    \expandafter\SF@gobble@opt
  \fi
}
\expandafter\def\csname SF@gobble@opt \endcsname{\@ifnextchar[
  \SF@gobble@twobracket
  \@gobble
}
\edef\SF@gobble@opt{\noexpand\protect
  \expandafter\noexpand\csname SF@gobble@opt \endcsname}
\def\SF@gobble@twobracket[#1]#2{}
\providecommand{\tabularnewline}{\\}

\usepackage{cvpr}
\usepackage{times}
\usepackage{epsfig}
\usepackage{graphicx}
\usepackage{amsmath}
\usepackage{amssymb}

\usepackage[pagebackref=true,breaklinks=true,letterpaper=true,colorlinks,bookmarks=false]{hyperref}

\cvprfinalcopy 

\def\cvprPaperID{122} 

\ifcvprfinal\fi

\makeatother

\usepackage{babel}

\usepackage{hyperref}
\usepackage{color}
\hypersetup{
  colorlinks=true,
  urlcolor=blue
}

\makeatother

\makeatletter
\def\and{%
  \end{tabular}%
  \hskip 0.5em \@plus.17fil\relax
  \begin{tabular}[t]{c}}
\makeatother

\usepackage{babel}
\begin{document}

\title{Augmenting Bag-of-Words: Data-Driven Discovery of Temporal and Structural
Information for Activity Recognition}

\author{
    Vinay Bettadapura$^{\text{1}}$\\
    \texttt{\small vinay@gatech.edu}
  \and
    Grant Schindler$^{\text{1}}$\\
    \texttt{\small schindler@gatech.edu}
  \and
    Thomas Plötz$^{\text{2}}$\\
	\texttt{\small thomas.ploetz@ncl.ac.uk}
  \and
    Irfan Essa$^{\text{1}}$\\
    \texttt{\small irfan@cc.gatech.edu}
  \and
    {\small $^{\text{1}}$Georgia Institute of Technology, Atlanta, GA, USA}
    \hspace{0.5em}
    {\small $^{\text{2}}$Newcastle University, Newcastle upon Tyne, UK}
  \and
    \href{http://www.cc.gatech.edu/cpl/projects/abow}{\small http://www.cc.gatech.edu/cpl/projects/abow}
}


\maketitle

\begin{abstract}
We present data-driven techniques to augment Bag of Words (BoW) models,
which allow for more robust modeling and recognition of complex long-term
activities, especially when the structure and topology of the activities
are not known a priori. Our approach specifically addresses the limitations
of standard BoW approaches, which fail to represent the underlying
temporal and causal information that is inherent in activity streams.
In addition, we also propose the use of randomly sampled regular expressions
to discover and encode patterns in activities. We demonstrate the
effectiveness of our approach in experimental evaluations where we
successfully recognize activities and detect anomalies in four complex
datasets.
\end{abstract}

\section{\label{sec:Introduction}Introduction}

Activity recognition in large, complex datasets has become an increasingly
important problem. Extracting activity information from time-varying
data has applications in domains such as video understanding, activity
monitoring for healthcare and surveillance. Traditionally, sequential
models like Hidden Markov Models (HMMs) and Dynamic Bayesian Networks
have been used to address activity recognition as a time-series analysis
problem. However, the assumption of Markovian dynamics restricts the
application of such sequential models to relatively simple problems
with known spatial and temporal structure of the data to be analyzed
\cite{ActivityRecognition-Survey}. Similarly, syntactic methods like
Parse Trees and Stochastic Context Free Grammars \cite{Moore-SCFG,Bobick-Ivanov}
are not well suited for recognizing weakly structured activities and
are not robust to erroneous or uncertain data. 

As a promising alternative, research in activity recognition from
videos and other time-series data has moved towards bag-of-words (BoW)
approaches and away from the traditional sequential and syntactic
models. However, while BoW approaches are good at building powerful
and sparser representations of the data, they completely ignore the
ordering and structural information of the particular words regarding
their absolute and relative positions. Furthermore, standard BoW approaches
do not account for the fact that different types of activities have
different temporal signatures. Each event in a long-term activity
has a temporal duration, and the time that passes between each pair
of consecutive events, is different for different activities 

We introduce novel BoW techniques and extensions that explicitly encode
the temporal and structural information gathered from the data. Recent
activity recognition approaches such as \cite{Niebles-IJCV-2008} have extended the BoW approach with topic
models \cite{wallach2006topic} using probabilistic Latent Semantic
Analysis \cite{pLSA-Hofmann} and Latent Dirichlet Allocation \cite{LDA-Blei},
leading to more complex classification methods built on top of standard
BoW representations. In contrast, we increase the richness of the
features in the BoW representation and with the use of standard classification
backends (like $k$-NN, HMM and SVM), we demonstrate that our augmented
BoW techniques lead to better recognition of complex activities.

\textbf{Contributions:} We describe a method to represent temporal
information by quantizing time and defining new temporal events in
a data-driven manner. We propose three encoding schemes that use $n$-grams
to augment BoW with the discovered temporal events in a way that preserves
the local structural information (relative word positions) in the
activity. This narrows the conceptual gap between BoW and sequential
models. In addition, to discover the global patterns in the data,
we augment our BoW models with randomly sampled Regular Expressions.
This sampling strategy is motivated by the random subspace method
as it is commonly used for decision tree construction \cite{bagging-predictors}
and related approaches which have shown success in a wide variety
of classification and visual recognition problems \cite{lepetit2006keypoint}.

We evaluate our approach in comparison to standard BoW representations
on four diverse classification tasks: \emph{i)} Vehicle activity recognition
from surveillance videos (Section \ref{sub:Ocean-City-Surveillance});
\emph{ii)} Surgical skill assessment from surgery videos (Section
\ref{sub:Surgery}); \emph{iii)} Unsupervised learning of player roles
in soccer videos (Section \ref{sub:Soccer}) and \emph{iv)} Recognition
of human behavior and anomaly detection in massive wide-area airborne
surveillance (simulation) data (Section \ref{sub:WAAS-Data}). Recognition
using our augmented BoW outperforms the standard BoW approaches in
all four datasets. We provide evidence that this superior performance
generalizes to any classification framework by demonstrating how sequential
models (HMMs), instance based learning ($k$-NNs), and discriminative
recognition techniques (SVMs) benefit from the new representation
and outperform respective models trained on standard BoW. Finally,
we show how augmented BoW-based techniques successfully unveil further
details of the analyzed datasets, such as behavior anomalies.

\section{Related Work}

The Bag of Words (BoW) model was first introduced for Information
Retrieval (IR) with text \cite{salton71}. Since then, it has been
used extensively for text analysis, indexing and retrieval \cite{IR:Stanford}.
Building on the success of BoW approaches for IR with text and images,
research in activity recognition has focused on working with BoW built
using local spatio-temporal features \cite{Wang-BMVC} and more recently
with robust descriptors, which exploit continuous object motion and
integrate it with distinctive appearance features \cite{AR-Chen},
features based on dense trajectories \cite{AR-Wang} and features
learnt in an unsupervised manner directly from video data \cite{AR-Le}.

While the focus has mostly been on recognizing human activities in
controlled settings, recent BoW based approaches have focused on recognizing
human activities in more realistic and diverse settings \cite{Laptev_b:learning},
and with the use of higher level semantic concepts (attributes) that
allow for more descriptive models of human activities \cite{attributes}.
However, when activities are represented as bags of words, the underlying
sequential information provided by the ordering of the words is typically
lost. To address this problem, $n$-grams have been used to retain
some of the ordering by forming sub-sequences of $n$ items \cite{IR:Stanford}
(Figure \ref{fig:Building-n-grams}). More recently, variants of the
$n$-gram approach have been used to represent activities in terms
of their local event sub-sequences \cite{Raffay:AI:2009}. While this
preserves local sequential information and causal ordering, adding
absolute and relative temporal information results in more powerful
representations as we demonstrate in this paper.

Our augmentation method is independent of the underlying BoW representation,
i.e., the modality of the data to be processed. The input to our algorithm
is a sequence of atomic events, i.e., words. On video data these can
be either derived from state-of-the-art short-duration event detectors
(e.g., the Actom Sequence Model \cite{actom-2011}, automatic action
annotation \cite{annotationVideos2009}), or any other suitable feature
detectors.

\section{Activity Recognition with Augmented BoW}

We define an \textit{activity} as a finite sequence of events over
a finite period of time where each \textit{event}
in the activity is an occurrence. For example, if ``start'', ``turn'',
``straight'' and ``stop'' are four individual events, then a vehicle
driving activity will be a finite sequence of those events over some
finite time (e.g. ``start $\rightarrow$ straight$\rightarrow$ turn$\rightarrow$
stop$\rightarrow$ start$\rightarrow$ straight$\rightarrow$ stop'').
We call these events\emph{, }that can be described by an observer
and have a semantic interpretation, as \emph{observable events}.

Recent methods for activity recognition try to detect such observable
events and build BoW upon it. However, the temporal structure underlying
the activities that shall be recognized is typically neglected. The
time taken by each observable event and the time elapsed between two
subsequent events are two important properties that contribute to
the temporal signature of an activity that is being performed. For
example, a car at a traffic light will have a shorter time gap between
the ``stop'' and ``start'' events than a delivery vehicle that
has to stop for a much longer time (until its contents are loaded/unloaded)
before it can start again.

\subsection{\label{sub:Temporal-Modeling}Discovering Temporal Information}

We represent activities as sequences of discrete, observable events.
Let $\omega=\{a_{1},a_{2},a_{3},\ldots,a_{p}\}$ denote a set of $p$
activities, and let $\phi=\{e_{1},e_{2},e_{3},\ldots,e_{q}\}$ denote
the set of $q$ types of observable events. Each activity $a_{i}$
is a sequence of elements from $\phi$. Each event type can occur
multiple times at different positions in $a_{i}$.

\begin{figure}
\begin{centering}
\includegraphics[width=0.5\columnwidth]{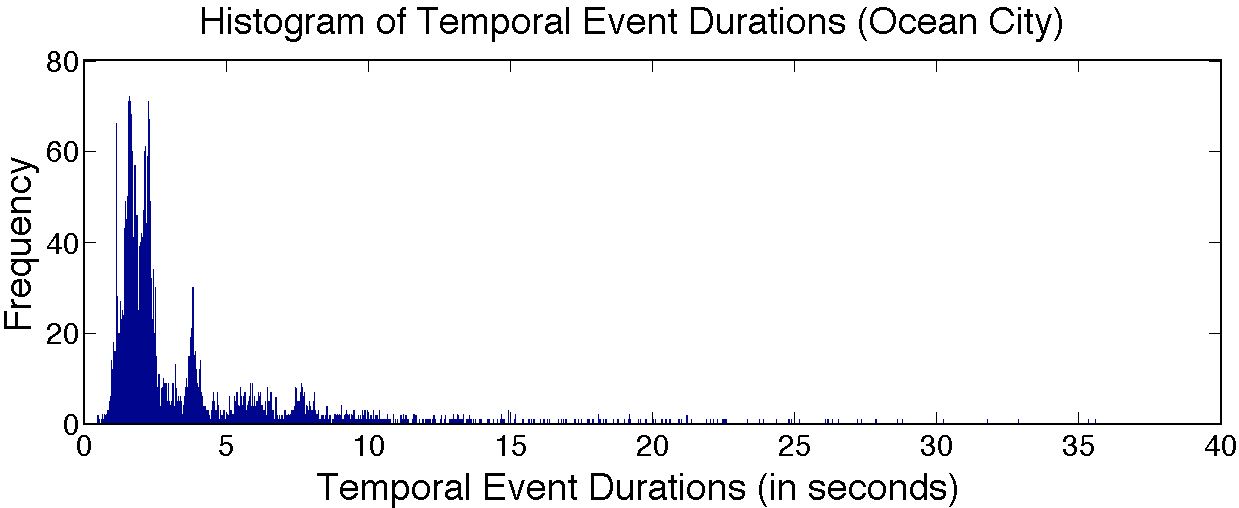}\includegraphics[width=0.5\columnwidth]{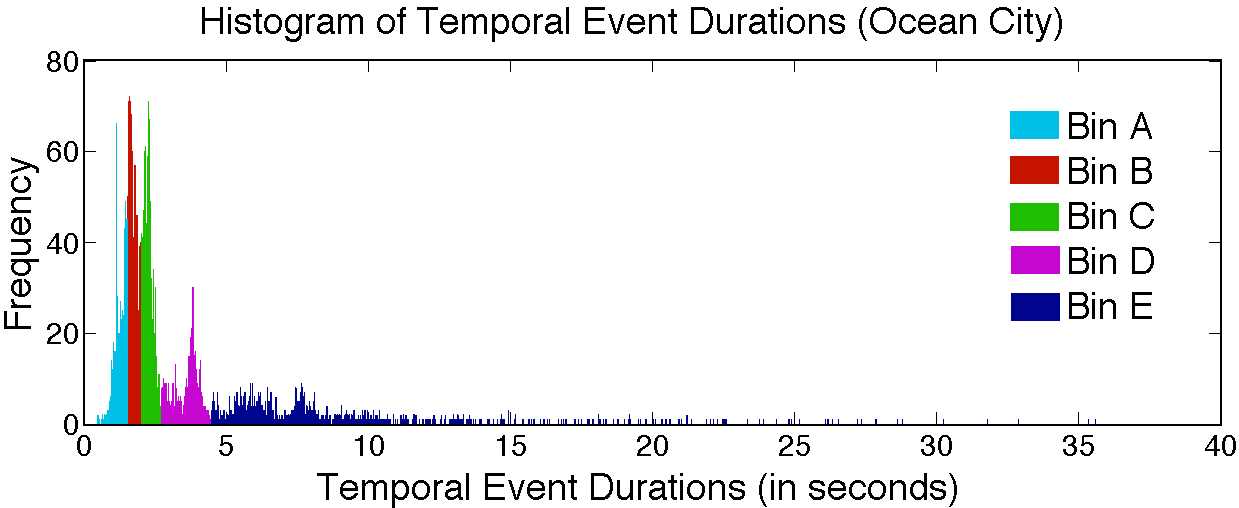}
\par\end{centering}

\caption{\label{fig:oc-time}Histogram of event durations for Ocean City dataset
(left) and \label{fig:oc-time-bins}data-driven creation of temporal
bins (right; $N=5$).}
\vspace{-1em}
\end{figure}

We now introduce \emph{temporal events}. Let $\tau_{j,k}$ be the
temporal event defined as the time elapsed between the end of observable
event $e_{j}$ and the start of observable event $e_{k}$, where $k>j$.
Since it measures time, $\tau_{j,k}$ is non-negative. Also, let $\pi_{j,k}$
be the temporal event defined as the time elapsed between the start
of observable event $e_{j}$ and the end of observable event $e_{k}$,
where $k\geq j$. Thus, $\tau_{j,k}$ measures the time elapsed between
any two events whereas $\pi_{j,k}$ measure the time elapsed between
any two events including the time taken by those two events. Thus,
$\tau_{j,k}$ and $\pi_{j,k}$ are related by the equation $\pi_{j,k}=\pi_{j,j}+\tau_{j,k}+\pi_{k,k}$.
We posit that these two types of temporal events, $\tau_{j,k}$ and
$\pi_{j,k}$, can model all the temporal properties of an activity.
The four possible scenarios are listed here:
\begin{enumerate}
\item $\tau_{j,j+1}$: Time elapsed between any two consecutive events $e_{j}$
and $e_{j+1}$
\item $\tau_{j,k}$: Time elapsed between any two events $e_{j}$ and $e_{k}$,
where $k>j$
\item $\pi_{j,j}$: Time taken by a single event $e_{j}$
\item $\pi{}_{j,k}$: Time taken by set of events $e_{j}$ to $e_{k}$,
where $k\geq j$
\end{enumerate}
To work with these temporal events, we will have to quantize them
into a finite number of $N$ bins. This quantization is crucial in
allowing us to incorporate a notion of time into BoW models. However,
uniformly dividing the time-line into $N$ bins is not ideal. As illustrated
by the temporal event duration histograms of $\tau_{j,j+1}$ for the
Ocean-City dataset (see Section \ref{sub:Ocean-City-Surveillance})
in Figure \ref{fig:oc-time}, short and medium duration temporal events
occur much more frequently than longer duration temporal events. Similar
temporal distributions are observed in the other datasets we have
analyzed.

To ensure that we capture the most useful temporal information, we
pursue a data-driven approach for binning. Bins are selected based
on the distribution of temporal events. If there are $S$ temporal
events, then we divide the temporal space into $N$ bins such that
each of the $N$ bins contains an equal proportion $S/N$ of the temporal
events (illustrated in Figure \ref{fig:oc-time-bins} for $N=5$).
Note that, if the time-line had been naively divided into $5$ equally
sized bins, then most of the temporal events would have
been placed in the first bin while the other $4$ bins would have
been almost empty. The choice of $N$ depends on the problem we are
addressing. Lower values of $N$ result in increased loss of temporal
information.

\textit{Example 1}: Say, temporal event $\tau_{j,k}$ is of $4$ second
duration and temporal event $\tau_{l,m}$ is of $20$ second duration,
then from Figure \ref{fig:oc-time-bins}, we see that $\tau_{j,k}$
will be assigned to bin \textbf{$D$ }and $\tau_{l,m}$\textbf{ }will
be assigned to bin $E$. Let $\psi$ denote the function that maps
the temporal events to their respective temporal bins. Then, we can
say that $\psi(\tau_{j,k})=D$ and $\psi(\tau_{l,m})=E$.

There are many possible ways by which we can encode these new temporal
events along with the observable events to build augmented BoW representations.
The simplest way would be to just add the quantized temporal events
to the BoW, i.e., if the BoW contained $x$ observable events and
we extracted $y$ new quantized temporal events, then the augmented
BoW will now contain $x+y$ number of elements. Although this naive
representation already gives better results than just the BoW (see
Section \ref{sec:Results}), as shown in the next section, more sophisticated
alternatives are possible.

\subsection{Encoding Local Structure}

In the following we describe three encoding schemes we have developed
that merge the temporal events with the observable events in a way
that captures local structure.

\subsubsection{\label{sub:Interspersed-Encoding}Interspersed Encoding}

In interspersed encoding, the main focus is on the time elapsed between
every pair of consecutive events. Let $\tau_{j,j+1}$ be a temporal
event defined as the time elapsed between any two consecutive observable
events $e_{j}$ and $e_{j+1}$ in activity $a_{i}$. Once the quantized
temporal events $\psi(\tau_{j,j+1})$ are computed for all event pairs
$e_{j},e_{j+1}\in a_{i}$, they are then inserted into $a_{i}$ at
their appropriate positions between events $e_{j}$ and $e_{j+1}$.
Let this new sequence of \textit{interspersed} events\emph{ }for activity
$a_{i}$ be denoted by $T_{i}$. In general, if activity $a_{i}$
has $d$ events, then after the inclusion of the quantized temporal
events, $T_{i}$ will have $2d-1$ events (the original $d$ observable
events plus the new $d-1$ temporal events).

\begin{figure}
\begin{centering}
\includegraphics[width=0.92\columnwidth]{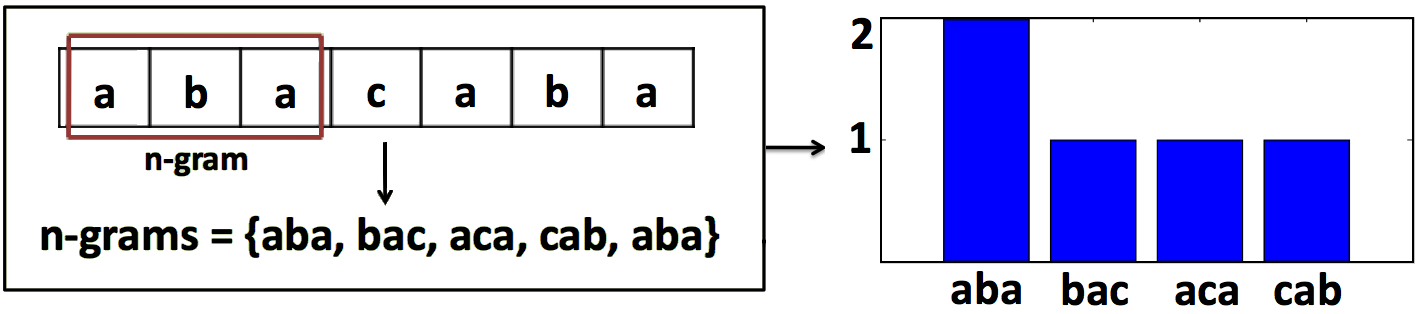}
\par\end{centering}

\caption{\label{fig:Building-n-grams}Building $n$-grams and their histogram
(here $n=3$) \cite{Raffay:AI:2009}}
\end{figure}

\textit{Example 2}: For the activity $a_{1}=(e_{1},e_{2},e_{3})$,
we have $T_{1}=(e_{1},\psi(\tau_{1,2}),e_{2},\psi(\tau_{2,3}),e_{3})$.
If temporal event $\tau_{1,2}$ is of $4$ second duration and $\tau_{2,3}$
is of $20$ second duration, then the quantized temporal events will
be $\psi(\tau_{1,2})=D$ and $\psi(\tau_{2,3})=E$. So, the interspersed
sequence of events for activity $a_{1}$, will be $T_{1=}(e_{1},D,e_{2},E,e_{3})$.

One of the main drawbacks of classical BoW representations is the
loss of original word orderings (i.e. local structural information).
This is particularly adverse in the context of activity recognition
because activities correspond to causal chains of observable and temporal
events. Losing the ordering will result in a loss of all causality
and contextual information. We employ $n$-grams in order to retain
ordering of events \cite{Fink2008-MMF}. An $n$-gram is a sub-sequence
of $n$ terms from a given sequence. Deriving $n$-grams and their
histograms from a given sequence is illustrated in Figure \ref{fig:Building-n-grams}.

Using this approach, for every activity $a_{i}$, the event sequence
$T_{i}$ is transformed into an $n$-gram sequence $T_{i}^{I}$ (where
the superscript $I$ stands for \textit{interspersed}). This $T_{i}^{I}$
feature vector representing activity $a_{i}$ is the final result
of interspersed encoding. From Example 2, with $n=3$, the event sequence
$T_{1=}(e_{1},D,e_{2},E,e_{3})$ will be transformed into the $n$-gram
sequence $T_{1}^{I}=(e_{1}De_{2},De_{2}E,e_{2}Ee_{3})$ or in its
histogram form $T_{1}^{I}=\{e_{1}De_{2}\Rightarrow1,De_{2}E\Rightarrow1,e_{2}Ee_{3}\Rightarrow1\}$
(denoted as key-value pairs where the key is the $n$-gram and the
value is its frequency).

\subsubsection{\label{sub:Cumulative-Encoding}Cumulative Encoding}

In cumulative encoding, the main focus is on the cumulative time taken
by a subsequence of observable events. Let $\psi(\pi_{j,j+n-1})$
be a quantized temporal event defined as the total time taken by $n$
consecutive events $e_{j}$ to $e_{j+n-1}$ in activity $a_{i}$.
Once the quantized temporal event $\psi(\pi_{j,j+n-1})$ is computed
for the consecutive sequence of observable events $e_{j}\ldots e_{j+n-1}\in a_{i}$,
it is appended to the set of the observable events. Let this new sequence
of ``cumulative'' observable and temporal\emph{ }events\emph{ }for
activity $a_{i}$ be denoted by $T_{i}^{C}$ (where the superscript
$C$ stands for ``cumulative'').

\textit{Example 3}: If activity $a_{2}=(e_{1},\ldots,e_{5})$, $n$
= 3, then $T_{2}^{C}=(e_{1}e_{2}e_{3}\psi(\pi_{1,3}),e_{2}e_{3}e_{4}\psi(\pi_{2,4}),e_{3}e_{4}e_{5}\psi(\pi_{3,5}))$.
Say, $\pi_{1,3}$ is of $4$ second duration, $\pi_{2,4}$ is of $20$
second duration and $\pi_{3,5}$ is of $1$ second duration and that
$\psi(\pi_{1,3})=D$, \textbf{$\psi(\pi_{2,4})=E$} and $\psi(\pi_{3,5})=A$.
So, the new sequence of events for activity $a_{2}$, will be $T_{2}^{C}=(e_{1}e_{2}e_{3}D,e_{2}e_{3}e_{4}E,e_{3}e_{4}e_{5}A)$
or in histogram form $T_{2}^{C}=\{e_{1}e_{2}e_{3}D\Rightarrow1,e_{2}e_{3}e_{4}E\Rightarrow1,e_{3}e_{4}e_{5}A\Rightarrow1\}$.

Interspersed encoding focuses on the time elapsed between events whereas
cumulative encoding focuses on the time taken by the events.

\subsubsection{\label{sub:Pyramid-Encoding}Pyramid Encoding}

Given the choice of encoding scheme ---either interspersed or cumulative---
in pyramid encoding all $l$-grams of length $l,\forall l\in\left[1,n\right]$
are generated. Then we build a pyramid of these $l$-grams allowing
for processing of event sequences at multiple scales of resolution.
We denote BoW representations for activity $a_{i}$ generated through
pyramid encoding by $T_{i}^{P}$.

The output of each of these encoding schemes, i.e., $T_{i}^{I}$,
$T_{i}^{C}$ and $T_{i}^{P}$ is the augmented BoW model containing
the observable and temporal events, encoded in a way that captures
the local structure.

\subsection{Capturing Global Structure}

While $n$-grams are good at capturing local information, their
capability to capture longer range relationships are rather limited.
This is where regular expressions come into play. Obviously, it is
computationally intractable to enumerate all possible regular expressions
for a given vocabulary of observable and temporal events. Thus, given
the set of observable events $\phi$ and the set of discovered temporal
events $N$, we construct a vocabulary of all events $\phi\cup N$
denoted by $\Gamma$ where $\left|\Gamma\right|=\left|\phi\right|+\left|N\right|$,
and create a sub-space of regular expressions by restricting their
form to:

\begin{equation}
\wedge\mbox{ }.*\mbox{ }(\alpha)\mbox{ }(\beta_{1}\mid\ldots\mid\beta_{r})\varphi\mbox{ }(\gamma)\mbox{ }.*\mbox{ }\$
\end{equation}
where the symbols $\alpha,\beta_{i},\gamma\in\Gamma$ with $i\in\left[1,r\right]$
and $r=\mbox{rand}\left(1,\left|\Gamma\right|\right)$. The symbol
$\varphi$ is randomly set to one of the three quantifier characters:
$\left\{ *,+,?\right\} $. The special characters have the following
meaning: $``\wedge"$ matches the start of the sequence, $``."$ matches
any element in the sequence, $``*"$ matches the preceding element
zero or more times, $``+"$ matches the preceding element one or more
times, $``?"$ matches the preceding element zero or one time and
$``\$"$ matches the end of the sequence. The $``|"$ operator matches
either of its arguments. For example, $e_{1}(e_{2}|e_{3})e_{4}$ will
match either $e_{1}e_{2}e_{4}$ or $e_{1}e_{3}e_{4}$.

The first symbol that will be matched $(\alpha)$ and last symbol
that will be matched $(\gamma)$ are chosen randomly from $\Gamma$
using probability-proportional-to-size sampling (PPS) and the $r$
intermediate symbols $\beta_{i}$ are chosen randomly from $\Gamma$
using simple random sampling (SRS). PPS concentrates on frequently
occurring events and picks the first and last symbols in the regular
expression to be the ones that have the greatest impact on the population
estimates whereas SRS chooses each of the intermediate symbols with
equal probability, thus giving a fair chance for all events to equally
participate in the matching process. The results of our experimental
evaluation suggest that this combination of PPS-SRS sampling of the
regular expression subspace strikes the right balance between discovering
global patterns across activities and discovering the anomalous activities.

Regular expressions of the above form are randomly generated and those
that do not match at least one of the activities/event-sequences are
rejected. Accepted regular expressions are treated as new words and
added to our augmented BoW representation. This final representation
now contains automatically discovered temporal information and both
local and global structural information of the activities. Our experiments
show that increasing the number of words in BoW through randomly generated
regular expressions by just $20\%$ boosts the activity recognition
and anomaly detection results significantly (Section \ref{sec:Results}).

\subsection{Activity Recognition}

Activity recognition using augmented BoW is pursued in a straightforward
manner by feeding the time-series data in their novel representation
into statistical modeling backends. Note that there is in principle
no limitation on the kind of classification framework to be employed.
In Section \ref{sec:Results} we present results for instance based
learning ($k$-NN), sequential modeling (HMM), and discriminative
modeling (SVM).

Given videos or time-series data of activities, temporal information
is discovered using the histogram method described in Section \ref{sub:Temporal-Modeling}.
Using $n$-grams, the temporal information is then merged with the
extracted BoW thereby preserving local ordering of the words. The
new BoW model is then further augmented by adding new words created
using randomly sampled regular expressions (to capture global patterns
in the data), and then processed by the statistical modeling backend
for actual activity recognition.

\section{\label{sec:Results}Experimental Evaluation}

The methods presented in this paper were developed in order to improve
BoW-based activity recognition, thereby aiming for generalization
across application domains. For practical validation, we have thus
evaluated our approaches in a range of experiments that cover three
diverse classes of learning problems (binary classification, multi-class
classification, and unsupervised learning) across four challenging
datasets from different domains.

\begin{figure}
\begin{centering}
\includegraphics[width=0.65\columnwidth]{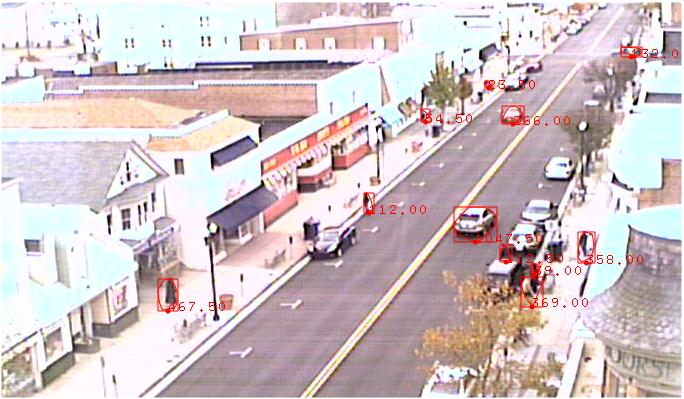}
\par\end{centering}

\caption{\label{fig:ocean-city-frame}Sample frame from Ocean City data showing
the various objects being tracked.}
\vspace{-1em}
\end{figure}

Optimization of the estimation procedure for augmented BoW representations
involves the two main parameters in our system: $N$, the number of
temporal bins used for quantization and $n$, the size of the $n$-gram
used for encoding. Low values of $N$ and $n$ result in the loss
of temporal and structural information whereas high values can lead
to large BoW with very high dimensionality. The optimal values for
$N$ and $n$ are determined by standard grid search \cite{libsvm-chang}.
Within a user supplied interval, all grid points of ($N$,$n$) are
tested to find the combination that gives the highest accuracy. $50\%$
of the particular datasets is held-out for parameter optimization,
and the remaining $50\%$ is used for model estimation using cross-validation.
This provides an unbiased estimate of the generalization error and
prevents over-fitting.

The main evaluation criterion for all activity recognition experiments
is classification accuracy, which we report as absolute percentages
and, for more detailed analysis, in confusion matrices. For the first
set of experiments (Section \ref{sub:Ocean-City-Surveillance}) we
compare three different classification backends ($k$-NNs with cosine-similarity
distance metric, HMMs, and SVMs) and explore their capabilities in
systematic evaluations of their parameter spaces. Due to space constraints
the presentation of results for the remaining set of experiments is
limited to those achieved with the $k$-NN classification backend.
These results are, however, representative for all three types of
classifiers evaluated.

$k$-NNs with cosine-similarity distance metric, i.e. Vector Space
Models (VSM), treat the derived BoW vectors of activities as document
vectors and allow for automatic analysis in terms of querying, classifying,
and clustering the activities \cite{IR:Stanford}. Prior to classification,
each term in our augmented BoW is assigned a weight based on its term-frequency
and document-frequency in order to obtain a statistical measure of
its importance. Classification is done using leave-one-out cross-validation
(LOOCV).

HMM-based experiments employ semi-continuous modeling with Gaussian
mixture models (GMM) as feature space representations \cite{Fink2008-MMF}.
GMMs are derived by means of an unsupervised density learning procedure.
All HMMs are based on linear left-right topologies with automatically
derived model lengths (based on training data statistics), and are
trained using classical Baum-Welch training. Classification is pursued
using Viterbi-decoding. Parameter estimation and model evaluation
employs $10$-fold cross-validation.

Experiments on SVMs are carried out in 10-fold cross-validation using
\textsc{LibSvm} with an RBF kernel. Parameter optimization utilizes
a grid-search procedure as it is standard for finding optimal values
for $C$ and $\lyxmathsym{\textgreek{g}}$ \cite{libsvm-chang}.

\subsection{\label{sub:Ocean-City-Surveillance}Ocean City Surveillance Data}

The first dataset consists of $7$ days of uncontrolled videos recorded
at Ocean City, USA \cite{Oh:2010:CRF:1886063.1886105}. The input
video was stabilized and geo-registered and $2,140$ vehicle tracks
were extracted using background subtraction and multi-object tracking
\cite{Oh:2010:CRF:1886063.1886105} (Figure \ref{fig:ocean-city-frame}).
An event detector analyzed the tracks, detected changes in structure
over time and represented each track by a sequence of observable events.
The types of events detected in each track were ``start'', ``stop'',
``turn'' and ``u-turn''.

\begin{figure}
\begin{centering}
\includegraphics[width=0.65\columnwidth]{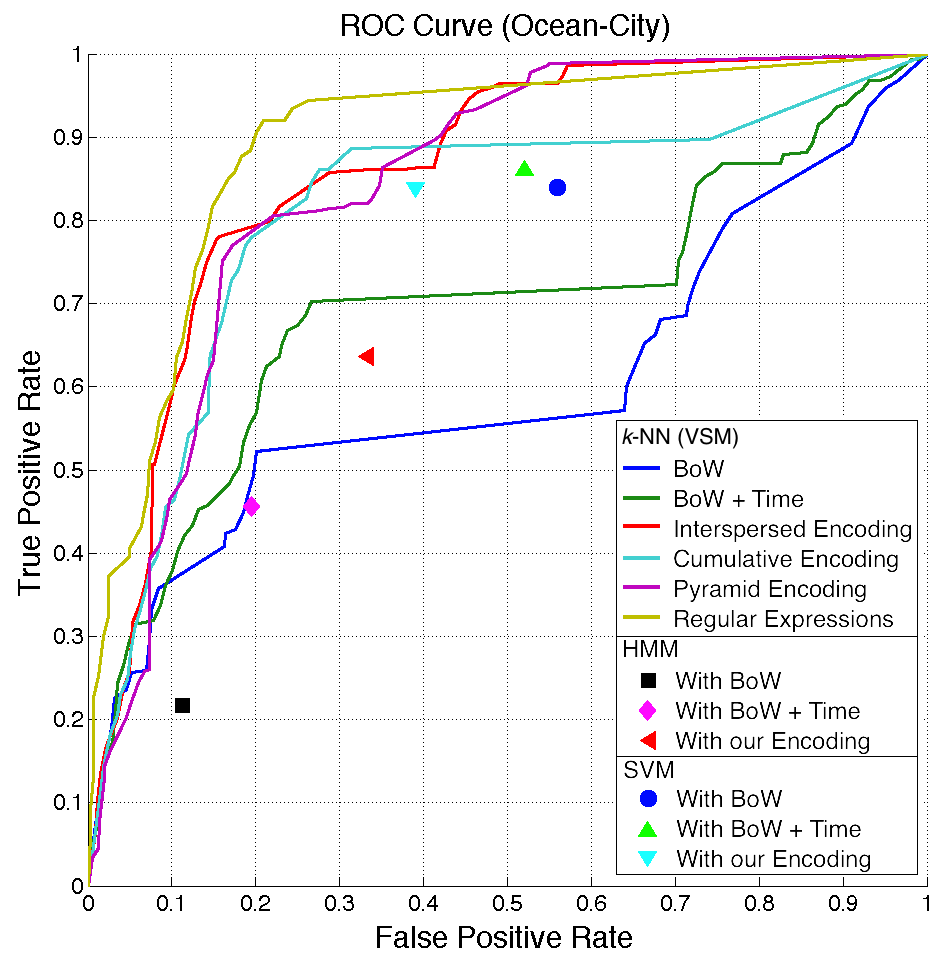}
\par\end{centering}

\caption{\label{fig:ocean-result}Classification results for Ocean City dataset.
Our encoding schemes outperform the BoW baseline on three classification
backends: VSM, sequential models (HMMs) and SVMs.}
\vspace{-1em}
\end{figure}

Out of the $2,140$ vehicle tracks, $448$ vehicles are either entering
or exiting parking areas on either side of the road (Figure \ref{fig:ocean-city-frame}).
The recognition objective is to determine whether or not vehicles
are involved in parking activities. 

With the empirically determined optimal values of $N=2$ and $n=2$,
we perform binary classification. The results are shown in Figure
\ref{fig:ocean-result}. For $k$-NN based experiments, ROC curves
were generated by varying the acceptance threshold. Augmenting BoW
with temporal information (bag-of-words + time) improves the results
over the BoW baseline. The performance is further improved with our
proposed Interspersed, Cumulative and Pyramid encoding schemes. However,
the best results are obtained when we augment our BoW with randomly
generated regular expressions.

Figure \ref{fig:ocean-result} also shows the performance of HMM and
SVM based recognition backends using augmented BoW representations.
Both techniques produce fixed decisions based on maximizing models'
posterior probabilities, i.e., no threshold-based post-processing
is applied for the actual recognition. Consequently, ROC curves are
not applicable, and the particular results are shown as points in
the figure.

\begin{table*}
\begin{centering}
\begin{tabular}{|l|>{\centering}p{0.06\paperwidth}|>{\centering}p{0.06\paperwidth}|>{\centering}p{0.06\paperwidth}|>{\centering}p{0.06\paperwidth}|>{\centering}p{0.06\paperwidth}|>{\centering}p{0.08\paperwidth}|>{\centering}p{0.08\paperwidth}|>{\centering}p{0.06\paperwidth}|}
\hline 
 & {\small Respect for tissue} & {\small Time and motion} & {\small Instrument handling} & {\small Suture handling} & {\small Flow of operation} & {\small Knowledge of procedure} & {\small Overall performance} & {\small Average accuracy}\tabularnewline
\hline 
\hline 
{\small M1: BOW baseline} & 66.67\% & 50.79\% & 50.79\% & 69.84\% & 49.21\% & 60.32\% & 52.38\% & 57.14\%\tabularnewline
\hline 
{\small M2: BOW + Time} & 69.84\% & 66.67\% & 65.08\% & 69.84\% & 63.49\% & 74.60\% & 68.25\% & 68.25\%\tabularnewline
\hline 
{\small M3: Our encoding} & \textbf{73.02\%} & \textbf{74.60\%} & \textbf{68.25\%} & \textbf{73.02\%} & \textbf{66.67\%} & \textbf{80.95\%} & \textbf{71.43\%} & \textbf{72.56\%}\tabularnewline
\hline 
\end{tabular}
\par\end{centering}

\medskip{}

\caption{\label{tab:Surgery-Results}Surgical skill assessment using OSATS
assessment scheme \cite{28martin1997objective}. Ground truth annotation
provided by an expert surgeon who assessed the training sessions using
7 different metrics (columns) and a three-point scale (low competence,
medium, and high skill). Results given are accuracies from automatic
recognition using $k$-NN, replicating expert assessment based on
video footage of the training sessions. Our encoding (Interspersed
encoding with $3$-grams, $5$ time bins and with $20$ random regular
expressions) outperforms the BoW baseline on all 7 metrics.}
\end{table*}

\begin{figure}
\begin{centering}
\includegraphics[width=0.49\columnwidth]{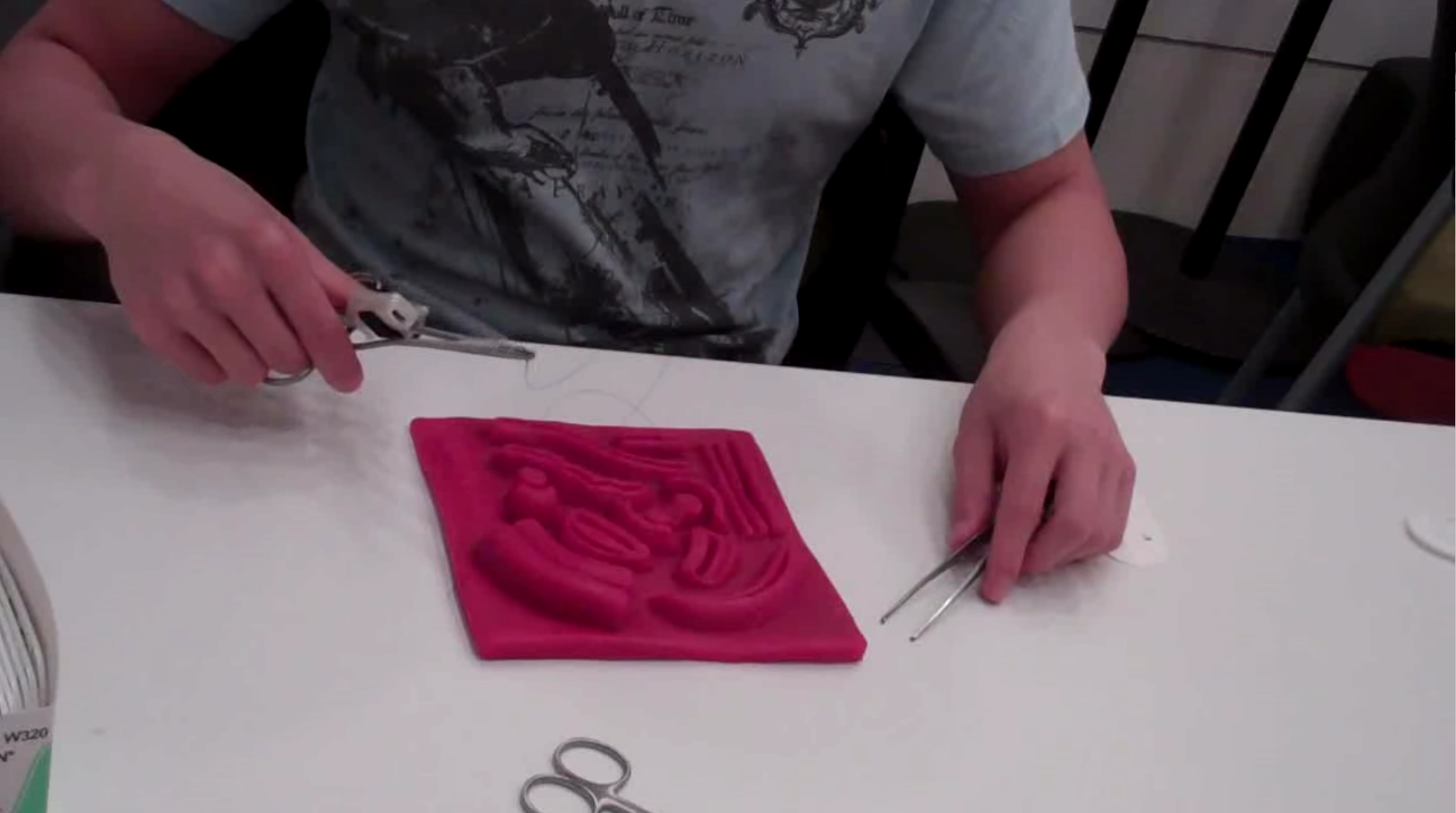}\ \ \includegraphics[width=0.49\columnwidth]{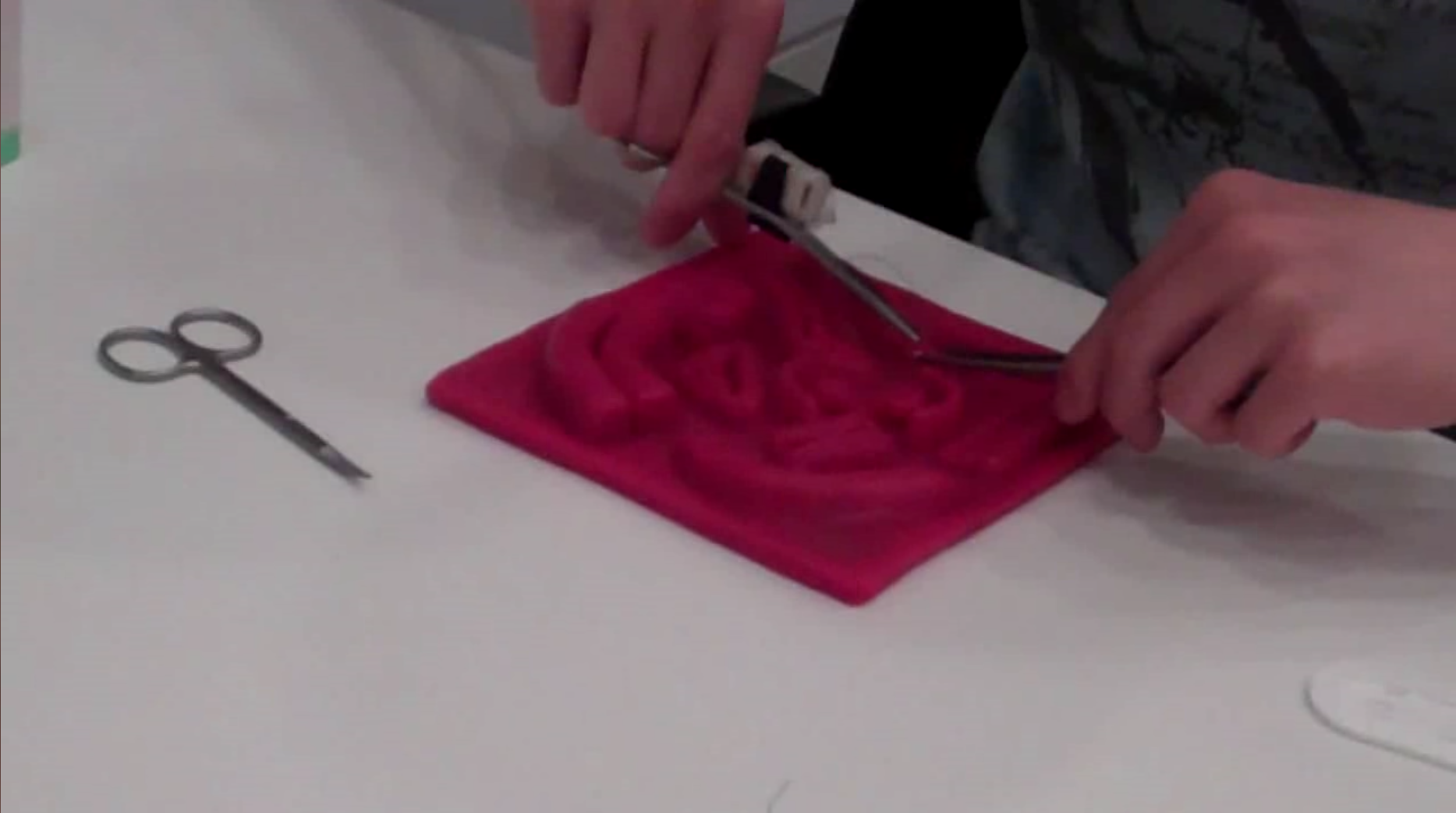}
\par\end{centering}

\caption{\label{fig:Surgery-Figs}Long-range (left) and close-up (right) stills
of video footage from training sessions for surgical skill assessment.
Participants practice suturing using regular instruments and suture
pads.}
\end{figure}

Analyzing the evaluation results, it becomes evident that: \emph{i)}
our proposed encoding schemes outperform the BoW baseline; and\emph{
ii)} superior classification accuracy generalizes across recognition
approaches ($k$-NN, HMM, SVM), with largest gain for Vector Space
Models.

\subsection{\label{sub:Surgery}Surgical Skill Assessment}

The second set of experiments is related to evaluating surgical skills
as it is standard routine in practical training of medical students.
As part of a larger case-study, $16$ medical students were recruited to
perform typical suturing activities (stitching, knot tying, etc.)
using regular instruments and tissue suture pads. Both long-range
and close-up videos of these ``suturing'' procedures were captured
at $50$ fps at a resolution of $720$p (sample still images in Figure
\ref{fig:Surgery-Figs}). As part of the training procedure, participants
completed $2$ sessions with $2$ attempts in each session, resulting
in a total of $64$ videos. Ground truth annotation was done by an
expert surgeon who assessed the skills of the participants using a
standardized assessment scheme (OSATS \cite{28martin1997objective})
based on $7$ different metrics (Table \ref{tab:Surgery-Results})
on a three-point scale (low competence, medium, and high skill).

Harris3D detectors and histogram of optical-flow (HOF) descriptors
\cite{Wang-BMVC} are used to extract visual-words from the surgery
videos. BoW are built with vocabularies constructed using $k$-means
clustering (with $k=50$), and then augmented using our techniques.
Table \ref{tab:Surgery-Results} summarizes our experiments (using
$k$-NN classification backend) and gives comparisons with the BoW
baseline. It can be seen that augmented BoW based approaches outperform
the BoW baseline in all $7$ skill metrics with an overall accuracy
of 72.56\%.

\begin{figure}
\begin{centering}
\includegraphics[width=0.49\columnwidth]{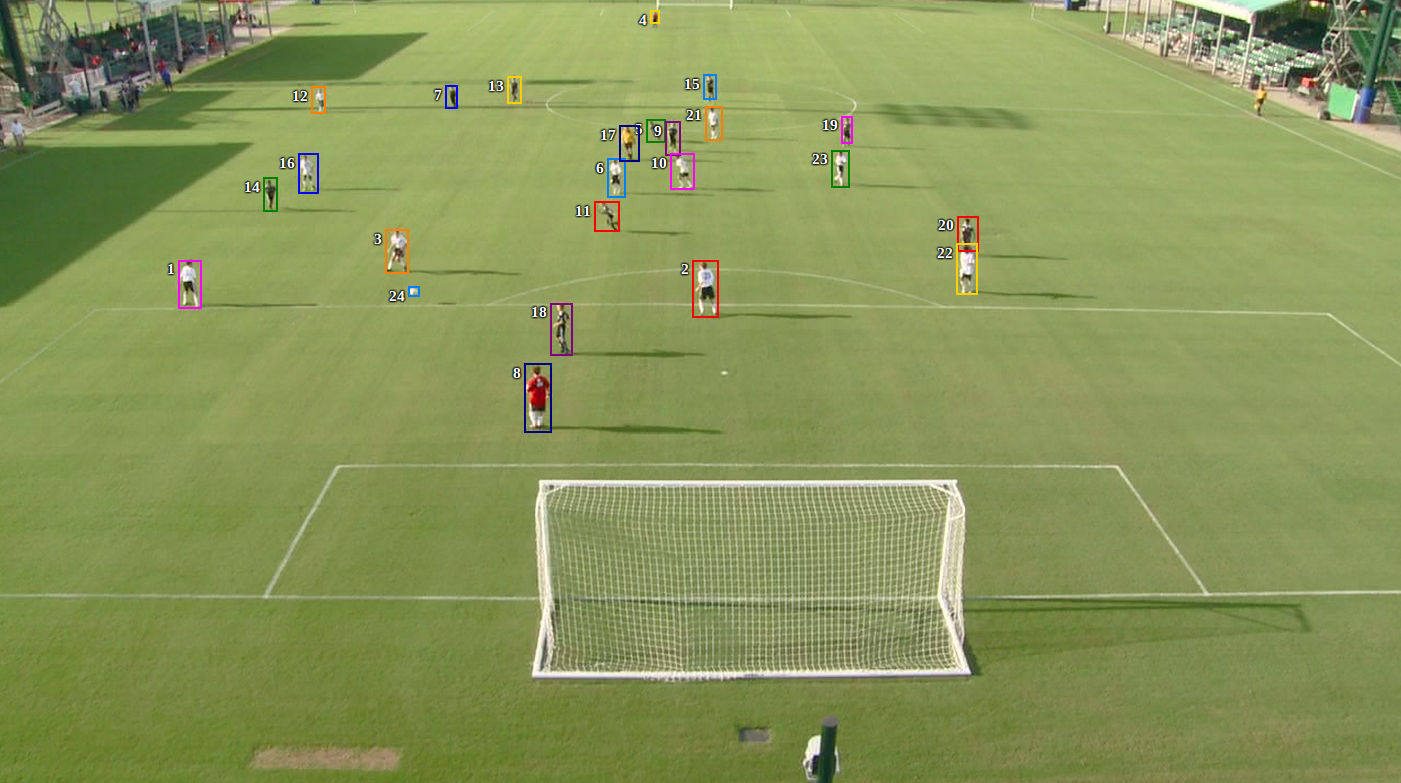}\ \ \includegraphics[width=0.49\columnwidth]{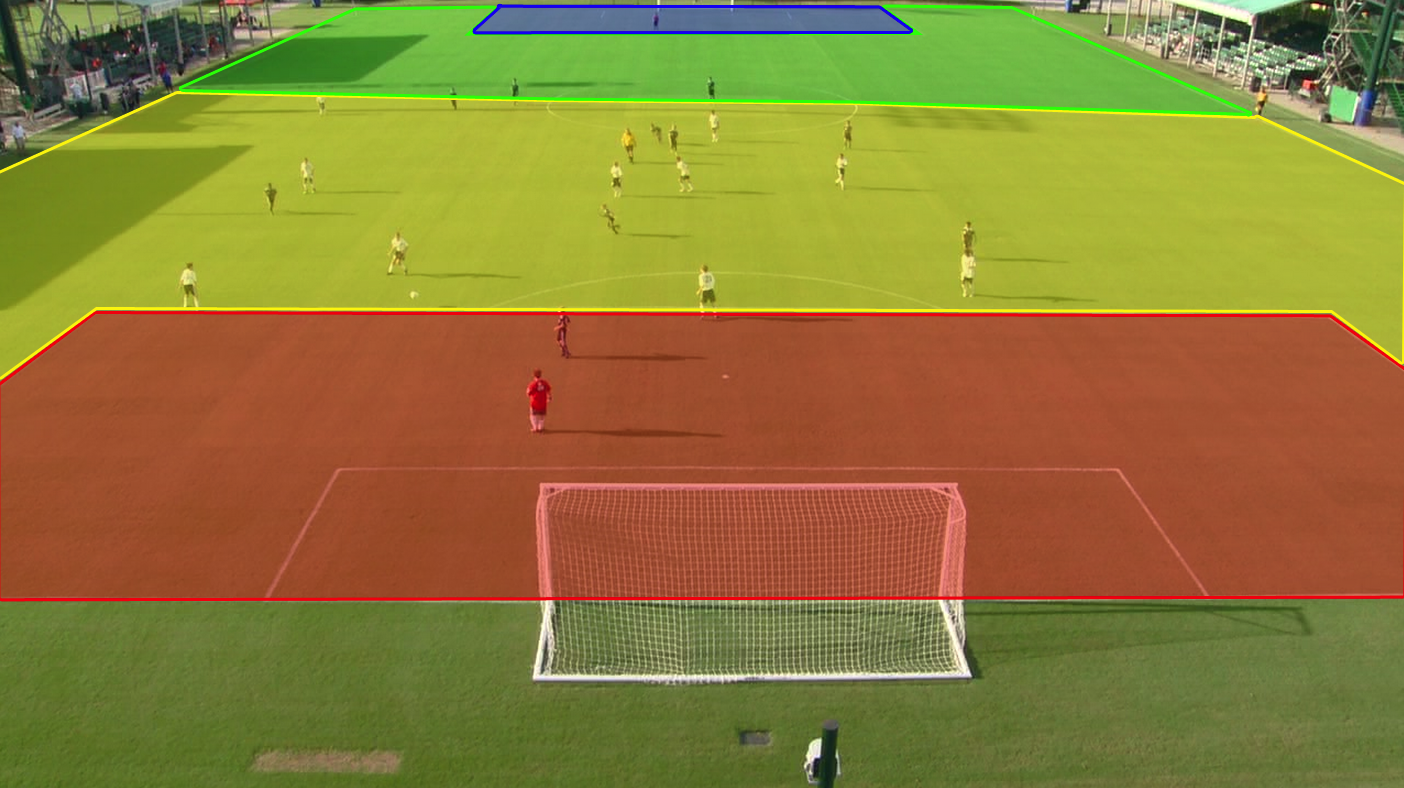}
\par\end{centering}

\caption{\label{fig:Soccer-Figs}Sample stills from soccer videos dataset.
\textbf{Left: }The $24$ objects being tracked: $22$ players from
both teams, referee and the ball. \textbf{Right:} The $4$ zones used
by our event detector: Zone-A (Red), Zone-B (Yellow), Zone-C (Green)
and Zone-D (Blue).}

\vspace{-0.5em}
\end{figure}

Since our augmented BoW representations capture time and co-occurrence
of words, we hypothesized that an automated analysis procedure using
augmented BoW should perform particularly well in assessing the ``time
and motion'' and ``knowledge of procedure'' skills. Recognition
results reported in Table \ref{tab:Surgery-Results} indicate that
this is indeed the case. The classification accuracies are 74.60\%
and 80.95\% (an increase of 23.81\% and 20.63\% respectively, over
the BoW baseline), thus validating our hypothesis.

\begin{table}[t]
\begin{centering}
\begin{tabular}{|>{\raggedright}p{0.34\columnwidth}|>{\centering}p{0.1\columnwidth}|>{\centering}p{0.1\columnwidth}|>{\centering}p{0.1\columnwidth}|}
\hline 
 & {\small RI} & {\small ARI} & {\small NMI}\tabularnewline
\hline 
\hline 
{\small BOW baseline} & {\small 0.7984} & {\small 0.2922} & {\small 0.6147}\tabularnewline
\hline 
{\small BOW + Time} & {\small 0.8300} & {\small 0.3920} & {\small 0.6974}\tabularnewline
\hline 
{\small Our Encoding} & \textbf{\small 0.8261} & \textbf{\small 0.5244} & \textbf{\small 0.7462}\tabularnewline
\hline 
\end{tabular}
\par\end{centering}

\medskip{}

\caption{\label{tab:Soccer-Results}Cluster quality on soccer videos dataset.
The 3 metrics used are Rand Index (RI), Adjusted Rand Index (ARI)
and Normalized Mutual Information (NMI). Our encoding (Interspersed
encoding with $3$-grams, $3$ time bins and 20 random regular expressions)
gives better cluster quality than the BoW baseline.}

\vspace{-1em}
\end{table}

\subsection{\label{sub:Soccer}Learning Player Activities from Soccer Videos}

Automatic detection, tracking and labeling of the players in soccer
videos is critical for analyzing team tactics and player activities.
Previous work in this area has mostly focussed on detecting and tracking
the players, recognizing the team of the players using appearance
models and detecting short-duration player actions. In our experiments,
we consider the problem of unsupervised learning of long-range activities
and roles the various players take on the field. Given their tracks,
we cluster them into $7$ clusters: ``Team-A-Goalkeeper'', ``Team-A-Striker'',
``Team-A-Defense'', ``Team-B-Goalkeeper'', ``Team-B-Striker'',
``Team-B-Defense'' and ``Referee''.

\begin{figure*}
\begin{centering}
\includegraphics[width=0.25\textwidth]{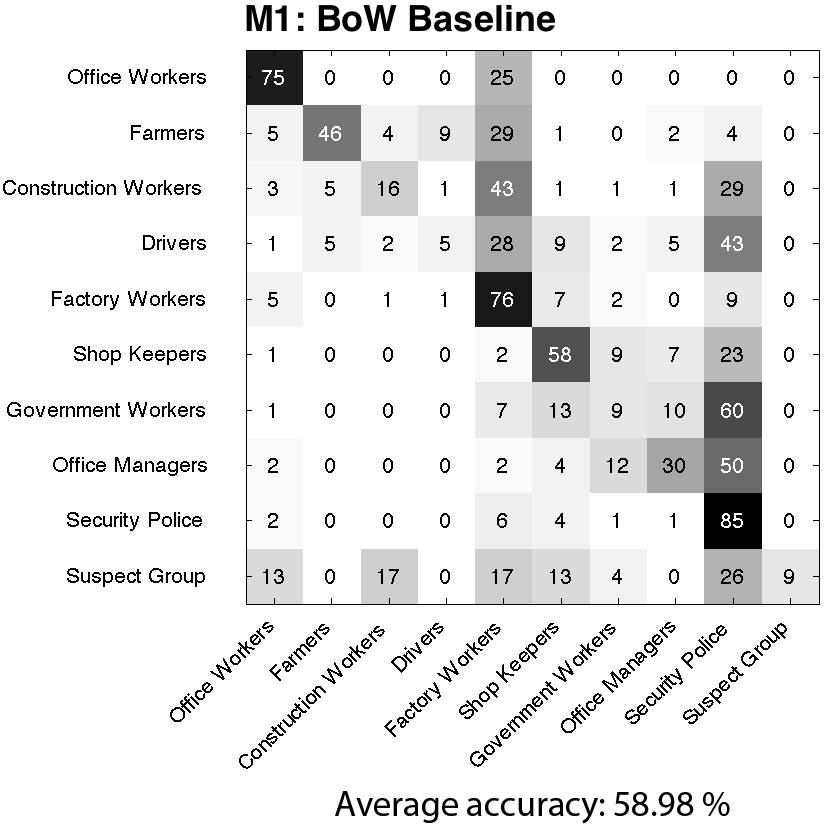}\ \ \ \ \ \ \includegraphics[width=0.25\textwidth]{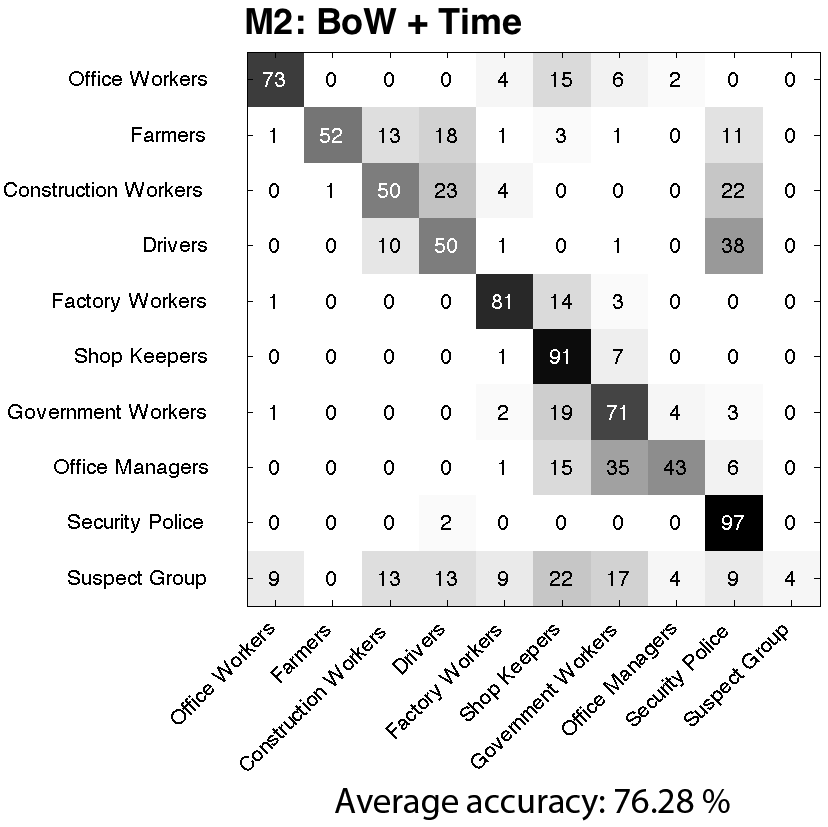}\ \ \ \ \ \ \includegraphics[width=0.25\textwidth]{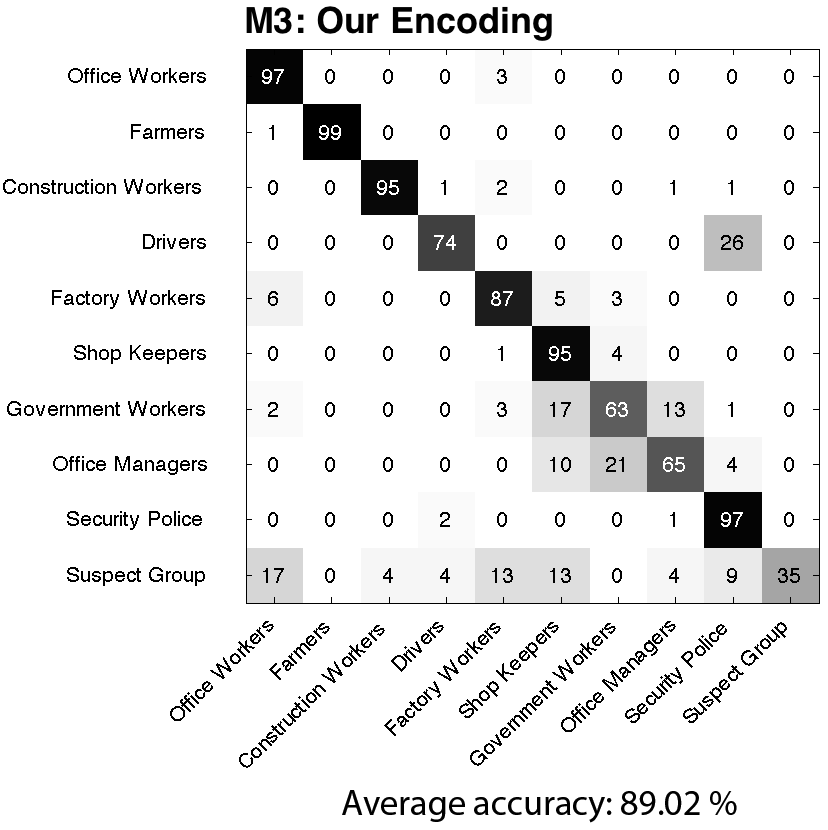}
\par\end{centering}

\caption{\label{fig:WAAS}Results on WAAS dataset: \textbf{Left:} BoW baseline;\textbf{
Middle: }BoW + Time; \textbf{Right: }Our encoding ($5$-grams, $5$
time bins and with $1,000$ random regular expressions). Overall improvement
of $30.04\%$ is observed with our method (compared to standard BoW
baseline).}
\end{figure*}

We analyzed full length match videos ($720$p at $59.94$ fps) from
the Disney Research soccer games dataset and tracked the $24$ objects
(players, referee, and ball) on the field using a multi-agent particle
filter based framework \cite{soccer-hamid} (Figure \ref{fig:Soccer-Figs}).
The tracks were given to an event detector that divided the field
into 4 zones (Figure \ref{fig:Soccer-Figs}) and detected $10$ types
of events: ``Enter-Zone-A'', ``Leave-Zone-A'', ``Enter-Zone-B'',
``Leave-Zone-B'', ``Enter-Zone-C'', ``Leave-Zone-C'', ``Enter-Zone-D'',
``Leave-Zone-D'', ``Receive-Ball'' and ``Send-Ball''. With this
vocabulary of $10$ events, we built augmented BoW and clustered them
using $k$-means clustering where $k=7$. Clustering results are
given in Table \ref{tab:Soccer-Results}. It can be seen that augmented
BoW outperform the BoW baseline on all $3$ cluster quality metrics.
In a supervised setting, we achieve an accuracy of $82.61\%$, which
is a $17.39\%$ improvement over the BoW baseline (which is $65.22\%$).

\subsection{\label{sub:WAAS-Data}Wide Area Airborne Surveillance (WAAS)}

In order to evaluate the applicability and scalability of our approach
on massive datasets with several hundreds of thousands of activities,
we consider the Wide-Area Airborne Surveillance (WAAS) simulation
dataset.

The WAAS dataset was developed by the U.S. Military as part of their
Activity Based Intelligence (ABI) initiative. The goal is to capture
motion imagery from an airborne platform that provides persistent
coverage of a wide area, such as a town or a small city, and merge
the automatically captured data from the aerial station with intelligence
gathered by ground forces to build a surveillance database of humans
and vehicles in that area. In order to aid research in this area,
the WAAS dataset has been released, which contains Monte Carlo simulation
of the activities of $4,623$ individuals for a total duration of
$46.5$ hours generated in $1$ minute increments. There are a total
of $180$ events (like ``Eat Lunch'', ``Enter Vehicle'', ``Exit
Vehicle'', ``Move'', ``Wait'', etc) with a total of $544,777$
event sequences spread across $28,682$ buildings. Ground truth labels
are available on the $10$ different professions of all the individuals.
$23$ out of the $4,623$ individuals are suspected to be part of
a terror group.

Given this large database, we show that our augmented BoW can successfully
classify people's professions and detect some of the suspect individuals
based on the temporal and structural similarities in their activities.
Classification accuracies and confusion matrices are shown in Figure
\ref{fig:WAAS}. Note that, with our encoding, more than a third of
the suspect group are correctly classified which baseline methods
failed to capture. This successful identification of suspicious behavior
is especially remarkable since those suspects aim for imitating \textquotedbl{}normal\textquotedbl{}
behavior and thus their activities are very similar to harmless activities.

\subsection{\label{sub:Statistical-Significance}Test for Statistical Significance}

With McNemar's chi-square test (with Yates' continuity correction),
we check for the statistical significance between the results of our
two multi-class classification problems (Figure \ref{fig:WAAS} and
Table \ref{tab:Surgery-Results}). For the surgery dataset, though
all the 7 skill classifications were statistically significant, due
to space constraints, only results on ``knowledge of procedure''
classification is presented. 

The null hypothesis is that the improvements are due to chance. However,
as shown in Table \ref{tab:McNemar's Test} for both the datasets,
the {\small $\chi^{2}$} values are greater than the critical value
(at 95\% significance level) of 3.84 and the $p$-values are less
than the significance level ($\alpha$) of 0.05. Thus, the null hypothesis
can be rejected and we can conclude that the improvements obtained
with our methods are statistically significant.

\begin{table}
\begin{centering}
{\scriptsize }%
\begin{tabular}{|c|c|c|}
\hline 
 & {\scriptsize M1 vs M2} & {\scriptsize M1 vs M3}\tabularnewline
\hline 
\hline 
{\scriptsize $\chi^{2}$} & {\scriptsize 165.09} & {\scriptsize 530.35}\tabularnewline
\hline 
{\scriptsize $p$-value} & {\scriptsize < 0.0001} & {\scriptsize 0.0026}\tabularnewline
\hline 
\end{tabular}{\scriptsize \ }%
\begin{tabular}{|c|c|c|}
\hline 
 & {\scriptsize M1 vs M2} & {\scriptsize M1 vs M3}\tabularnewline
\hline 
\hline 
{\scriptsize $\chi^{2}$} & {\scriptsize 4.76} & {\scriptsize 9.33}\tabularnewline
\hline 
{\scriptsize $p$-value} & {\scriptsize 0.0291} & {\scriptsize 0.0023}\tabularnewline
\hline 
\end{tabular}
\par\end{centering}{\scriptsize \par}

\medskip{}

\caption{\label{tab:McNemar's Test}McNemar's tests on statistical significance
between the different methods on the 2 multi-class classification
problems. Each column compares two methods. \textbf{Left: }Comparing
the methods in Figure \ref{fig:WAAS} for the WAAS dataset; \textbf{Right:}
Comparing the methods in Table \ref{tab:Surgery-Results} for the
``knowledge of procedure'' skill in the surgery dataset (the other
6 skill classifications were also statistically significant, but are
not shown due to space constraints).}
\vspace{-1em}
\end{table}

\section{Conclusion}

BoW models are a promising approach to real-world activity recognition
problems where only little is known a-priori about the underlying
structure of the data to be analyzed. We presented a significant extension
to BoW-based activity recognition, where we augment BoW with temporal
information and with both local and global structural information,
using temporal encoding, $n$-grams and randomly sampled regular expressions,
respectively. In addition to generally improved activity recognition,
our approach also detects anomalies in the data, which is important,
for example in human behavior analysis applications. We have demonstrated
the capabilities of our approach on both real-world vision problems
and on massive wide-area surveillance simulations.\textbf{}\\
\textbf{}\\
\textbf{Acknowledgements} Funding and sponsorship was provided by the U.S. Army Research Office (ARO) and Defense Advanced Research Projects Agency (DARPA) under Contract No. W911NF-11-C-0088 and W31P4Q-10-C-0262. Parts of this work have also been funded by the RCUK Research Hub on Social Inclusion through the Digital Economy (SiDE), and the German Research Foundation (DFG, Grant No. PL554/2-1). Any opinions, findings and conclusions or recommendations expressed in this material are those of the authors and do not necessarily reflect the views of our sponsors.

We thank Kitware, Disney Research and DARPA for providing the Ocean City, Soccer and WAAS datasets, respectively.

\small\bibliographystyle{plain}
\bibliography{vsm-cvpr-sorted}

\end{document}